\documentclass[sigconf, screen, urlbreakonhyphens, 9pt]{acmart}

\usepackage{multirow}
\usepackage{threeparttable}
\usepackage{subfig}
\usepackage{changepage}
\usepackage{enumitem}
\usepackage[ruled,vlined]{algorithm2e}

\definecolor{deepgreen}{rgb}{0.0, 0.5, 0.0}

\setlist[itemize]{leftmargin=*}

\renewcommand\footnotetextcopyrightpermission[1]{}
\begin{document}

\title{\emph{Deep-Lock}: Secure Authorization for Deep Neural Networks}
\subtitle{\small \textbf{NOTE}: An extended version of this work is published in the ACM Journal on Emerging Technologies in Computing Systems~\cite{alam2022nn}.}

\author{Manaar Alam}
\email{alam.manaar@gmail.com}
\affiliation{
  \institution{Indian Institute of Technology Kharagpur}
  \state{INDIA}
}

\author{Sayandeep Saha}
\email{sayandeep.iitkgp@gmail.com}
\affiliation{
  \institution{Indian Institute of Technology Kharagpur}
  \state{INDIA}
}

\author{Debdeep Mukhopadhyay}
\email{debdeep.mukhopadhyay@gmail.com}
\affiliation{
  \institution{Indian Institute of Technology Kharagpur}
  \state{INDIA}
}

\author{Sandip Kundu}
\email{kundu@umass.edu}
\affiliation{
  \institution{University of Massachusetts Amherst}
  \state{USA}
}

\begin{abstract}
Trained Deep Neural Network (DNN) models are considered valuable Intellectual Properties (IP) in several business models. Prevention of IP theft and unauthorized usage of such DNN models has been raised as of significant concern by industry. In this paper, we address the problem of preventing unauthorized usage of DNN models by proposing a generic and lightweight key-based model-locking scheme, which ensures that a locked model functions correctly only upon applying the correct secret key. The proposed scheme, known as \emph{Deep-Lock}, utilizes \textit{S-Box}es with good security properties to encrypt each parameter of a trained DNN model with secret keys generated from a master key via a key scheduling algorithm. The resulting dense network of encrypted weights is found robust against \textit{model fine-tuning attacks}. Finally, Deep-Lock does not require any intervention in the structure and training of the DNN models, making it applicable for all existing software and hardware implementations of DNN.

\end{abstract}


\maketitle
\pagestyle{plain}

\section{Introduction}
The remarkable success of Deep Neural Networks (DNN) has created numerous opportunities for commercial applications. Constructing accurate DNN models for targeted applications is non-trivial, as it demands domain expertise and powerful computing resources. On the other hand, the datasets used for training are deemed one of the most valuable assets of modern businesses. Consequently, modern business models consider trained DNNs as Intellectual Property (IP) cores. The presence of such IP cores both in embedded platforms (Google Coral, Intel NCS) and cloud-based Machine-Learning-as-a-Service (MLaaS) (Amazon AWS, BigML) creates multiple practical security concerns. One such potent threat is the stealing decision boundary of the IP Core (also called \emph{Model Stealing}), which has been addressed previously by several work~\cite{tramer2016stealing,yu2020cloudleak}. Businesses may suffer economic losses, as well as losses of brand value due to such incidents.

In this work, we address another threat model, which has been proposed recently in~\cite{dac_abhishek}. This is concerned with the \textit{model piracy} of DNN models that may also result in significant economic losses for the original model owner. While model stealing may also lead to illegal usage, the problem of model piracy is more fundamental. An adversary may obtain/access a model from several other unwanted sources and continue using it as a black box (without bothering stealing its decision boundary). Unless there is some mechanism implemented for authorized access, this threat can never be mitigated. While this threat is applicable both in the context of MLaaS and embedded platforms, it is more severe in the latter case as the model can be easily extracted as a piece of software and distributed among multiple parties.  

Although there exist techniques for watermarking a given DNN model for establishing model ownership~\cite{adi2018turning,guo2018watermarking,darvish2019deepsigns}, they cannot prevent its illegal usage. Hence, an explicit locking mechanism for DNN is in demand, which will prevent unauthorized usage by making the model malfunction. Previous research in this area has mainly focused on obfuscating model structures~\cite{xu2018deepobfuscation}. However, a more critical component of a DNN IP is the trained parameters. Most industrial applications typically use previously published DNN architectures, which have demonstrated high modeling capabilities but use different parameters depending on applications. Observing this, we present \emph{Deep-Lock}, which enables a secret key-based locking for DNN parameters. Unauthorized access, without the knowledge of the legitimate key, makes the model unusable by severely degrading the model accuracy. Deep-Lock utilizes S-Boxes with good cryptographic properties to lock each trained parameter of a DNN model, which is found to be a lightweight approach for model locking, without causing any significant degradation in the inference time and model size. The secret keys required for locking are generated from a master key via a key schedule. The master key is only needed to be stored within the device. Moreover, the locking mechanism works on trained models and has no interaction with the training phase, making it generic and scalable for different model types. We have verified our claims over several industry-scale DNN models. We also considered improving the locked model accuracy by \emph{model fine-tuning}, where a \emph{manifest dataset}\footnote{The manifest dataset is a set of labeled data that is apparent to the user, though not necessarily a subset of original training dataset, but resembles sufficiently to the original training dataset.} is used for improving the model accuracy even for wrong key values. It was found that model fine-tuning fails to improve a model's accuracy to a practically significant value.\vspace{-0.05cm}

Deep-Lock bears certain similarities with a key-dependent hardware-assisted DNN IP protection scheme (HPNN) proposed in~\cite{dac_abhishek}. However, HPNN requires changes to be made in the model structure, which is not there in Deep-Lock, easing its usability. Furthermore, Deep-Lock uses established cryptographic constructs for locking, which is lacking in HPNN.  It is worth noting that \textit{Deep-Lock} differs significantly from state-of-the-art software product key schemes, where a key is verified before giving access to a locked software. The security of such schemes depends on the verification step, and there exist several well-known techniques to bypass this verification check and gain control over the entire product. Furthermore, once the verification step is passed, the unlocked model remains in the memory from where it can be stolen without much hindrance. On the other hand, Deep-Lock embeds the verification mechanism inside the DNN and mandates verification for every query made to the network. This never leaves the unlocked model in the memory, making the direct steal of the model challenging.

The rest of the paper is organized as follows: Section~\ref{sec:threat_model} discusses the adversarial threat model considered in this work. Section~\ref{sec:methodology} presents the proposed methodology in details. The evaluation of performance and security of the proposed model is demonstrated in Section~\ref{sec:performance} and Section~\ref{sec:security_evaluation}, respectively. Finally, Section~\ref{sec:conclusion} concludes the work.\vspace{-0.1cm}

\section{Threat Model}\label{sec:threat_model}
The primary objective of this work is that a DNN model owner will provide services only to authorized customers who have acquired a license for model usage by paying an amount. To achieve the objective, the model owner encrypts the DNN model with a secret key. An adversary aims to use the encrypted DNN model accurately without paying a license fee to the model owner. In this scenario, we consider an adversary who has white-box access to the encrypted model, i.e., she knows model structure as well as all encrypted parameter values. The only information she does not know is the secret key used to encrypt the model. We also assume that the adversary has a \textit{manifest dataset}, which she can use to query the model with a key guess to obtain an output. If the adversary uses encrypted parameters in the stolen DNN architecture, it produces wrong input-output mapping as she does not have a legitimate secret key.\vspace{-0.1cm}

\section{Proposed Methodology}\label{sec:methodology}
The proposed key-based authorization scheme \textit{Deep-Lock} operates in two modes -- offline mode and online mode. In the offline mode, a DNN model owner trains and locks a model, and in the online mode, the DNN model is deployed for practical usage. The overview of all the operations performed in the \textbf{offline mode} is as follows:
\begin{itemize}
    \item First, a DNN model owner acquires a massive set of data related to a particular task and expends a considerable amount of money, experts knowledge, and computational resources to label the data and train a model. The model owner can exercise any popularly used learning technique for fine-tuning model parameters to obtain the most accurate model.
    \item The DNN model owner then selects a master key and uses a key-scheduling algorithm to generate different secret keys corresponding to each trained parameter. The parameters are then encrypted with an \textit{S-Box} operation using the derived secret keys. The original model parameters are replaced with these encrypted values before deploying it for practical usage. The lock operation is shown in Algorithm~\ref{algo:lock}.
    \item Finally, the DNN model owner distributes the locked model to authorized customers for practical usage. In a cloud-based service, the secret key can be distributed along with an access to the locked model. However, in a deployed hardware device, the secret key can be embedded into a trusted platform module (TPM), assuming that an adversary can neither access the TPM nor retrieve the key from it.
\end{itemize}

In the \textbf{online mode} of operation, a user needs to produce a master key during each query. The same key-scheduling algorithm used during the offline mode of operation takes the master key as input and generates decryption keys for each locked parameters. The parameters are then decrypted with the \textit{Inverse S-Box} operation and the derived secret keys. The unlock operation is shown in Algorithm~\ref{algo:unlock}. Thus, for a correct master key, all the locked parameters will be correctly decrypted, and the model will produce a correct prediction. However, for an incorrect key the model will produce a wrong prediction.

\begin{algorithm}[!t]
\SetAlgoLined
\SetKwInOut{Input}{Input}
\SetKwInOut{Output}{Output}
\Input{Set of $n$ real-valued trained DNN parameters: $\{w_0, w_1, \dots, w_{n-1}\}$; A master key $\mathcal{K}$; A $Key\_Schedule$ algorithm; A substitution-box mapping $Sbox$}
\Output{Locked DNN parameters: $\{w_0^{\prime}, w_1^{\prime}, \dots, w_{n-1}^{\prime}\}$}
$\{k_0, k_1, \dots, k_{n-1}\}$ = $Key\_Schedule(\mathcal{K})$\;
\For{$i\gets0$ \KwTo $n-1$}{
    $w_i^{bin}$ = binary representation of $w_i$\;
    $\mathcal{E}_{k_i}(w_i^{bin})$ = $Sbox[w_i^{bin} \oplus k_i]$\;
    $w_i^{\prime}$ = converted real value of $\mathcal{E}_{k_i}(w_i^{bin})$\;
}
\caption{Lock Operation\label{algo:lock}}
\end{algorithm}
\vspace{-0.25cm}
\begin{algorithm}[!t]
\SetAlgoLined
\SetKwInOut{Input}{Input}
\SetKwInOut{Output}{Output}
\Input{Set of $n$ real-valued locked DNN parameters: $\{w_0^{\prime}, w_1^{\prime}, \dots, w_{n-1}^{\prime}\}$; An input key $\mathcal{K}_{input}$; A $Key\_Schedule$ algorithm; An inverse substitution-box mapping $InvSbox$}
\Output{Unlocked DNN parameters: $\{w_0^{unlocked}, w_1^{unlocked}, \dots, w_{n-1}^{unlocked}\}$}
$\{k_0^{input}, k_1^{input}, \dots, k_{n-1}^{input}\}$ = $Key\_Schedule(\mathcal{K}_{input})$\;
\For{$i\gets0$ \KwTo $n-1$}{
    $(w_i^{bin})^{\prime}$ = binary representation of $w_i^{\prime}$\;
    $(w_i^{bin})^{unlocked}$ = $InvSbox[(w_i^{bin})^{\prime}] \oplus k_i^{input}$\;
    $w_i^{unlocked}$ = converted real value of $(w_i^{bin})^{unlocked}$\;
}
\caption{Unlock Operation\label{algo:unlock}}
\end{algorithm}

\section{Performance Evaluation}\label{sec:performance}
The method is evaluated on Intel Xeon CPU E5-2690 v4 @ 2.60GHz with 56 cores. The Convolutional Neural Network (CNN) structures and the number of trainable parameters for each evaluation dataset are shown in Table~\ref{table:dataset}.

\begin{table}[!t]
\centering
\caption{Datasets and Models\label{table:dataset}\vspace{-0.3cm}}
\begin{threeparttable}
\resizebox{0.75\linewidth}{!}{
\begin{tabular}{|c|c|c|}
\hline
\textbf{Dataset} & \textbf{Network Structure} & \textbf{\begin{tabular}[c]{@{}c@{}}Number of\\ Parameters\end{tabular}} \\ \hline
\textbf{MNIST} & C, MP, C, MP, FC, FC & 86,166 \\ \hline
\textbf{Fashion-MNIST} & C, MP, C, MP, FC, FC & 180,438 \\ \hline
\textbf{CIFAR-10} & C, C, MP, C, C, MP, FC, FC & 1,250,858 \\ \hline
\end{tabular}}
\begin{tablenotes}
\item \small \textbf{C:} Convolution Layer, \textbf{MP:} MaxPool Layer, \textbf{FC:} Fully Connected Layer\vspace{-0.6cm}
\end{tablenotes}
\end{threeparttable}
\end{table}

\subsection{Accuracy of Locked Models}
\begin{figure}[!t]
    \centering
    \subfloat[] {\label{fig:accuracy}
        \includegraphics[width=0.5\linewidth, height=3cm]{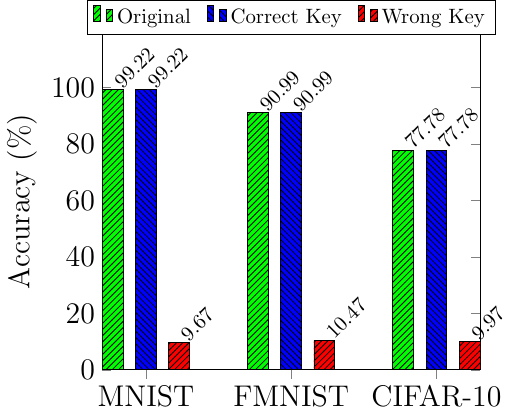}
    }
    \subfloat[] {\label{fig:timing_overhead}
        \includegraphics[width=0.5\linewidth, height=3cm]{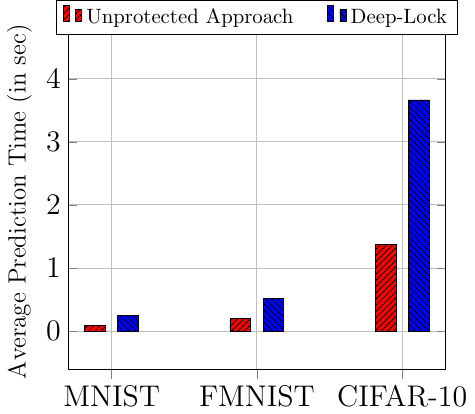}
    }
    \vspace{-0.4cm}
    \caption{(a) Classification accuracy of the original trained model, locked model with a correct key input, and locked model with a wrong key guess (b) Average prediction time of a single input for both unencrypted and locked model}
\end{figure}

We have locked each model mentioned in Table~\ref{table:dataset} with a master key using \textit{Deep-Lock}. Without loss of generality, we have considered AES SBox and AES key-scheduling algorithm to lock the models. Figure~\ref{fig:accuracy} shows the original classification accuracies of the trained CNN models, the classification accuracies of the locked models for correct master key, the average accuracies of the locked models over 100 randomly selected incorrect keys. \textit{We can observe that Deep-Lock does not compromise the accuracy of the trained models.} We can also observe that, for a wrong key, locked models work as a random classifier.

\subsection{Performance Overhead}
The average response time of both unencrypted and locked DNN models for classifying a single input are shown in Figure~\ref{fig:timing_overhead}. We can observe that the overhead is not significant for MNIST and Fashion-MNIST. However, we obtain a comparatively higher overhead for the CIFAR-10 dataset. The figures are based on the sequential execution of software implementations of the DNN models. However, these can be significantly improved with high level of parallelization in dedicated hardware accelerators.

\section{Security Evaluation}\label{sec:security_evaluation}
\begin{figure}[!t]
    \centering
	\includegraphics[width=0.5\linewidth]{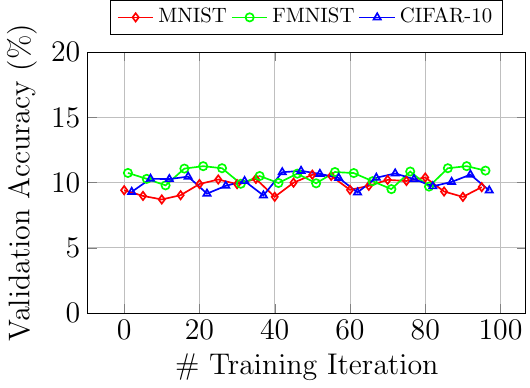}
	\vspace{-0.4cm}
    \caption{Validation accuracies for each DNN model over training iterations during model fine-tuning\label{fig:fine_tuning}}
\end{figure}

In order to compare the security benefits with the recent lightweight hardware-assisted key-dependent DNN obfuscation framework HPNN~\cite{dac_abhishek}, we provide a security evaluation of \textit{Deep-Lock} considering \textit{model fine-tuning attacks}. In a model fine-tuning attack, an adversary's objective is to retrain the locked model to an optimum parameter setting, which is different from the original, to obtain a comparative performance of the original model. The assumptions for an adversary in this attack are that she has the expertise and powerful computational resources to train any DNN model. To perform the model fine-tuning attack, an adversary obtains the DNN architecture and the model parameters from a locked model and utilizes the \textit{manifest dataset} to retrain the model. The attack is considered successful if the adversary obtains high accuracy with a wrong key from the locked model. For this experiment, we have considered that the adversary has access to 10\% of the training data. It is shown in~\cite{dac_abhishek} that with this dataset, an adversary can fine-tune a DNN model locked with the HPNN framework to achieve 82.43\% and 78.53\% accuracy for Fashion-MNIST and CIFAR-10 dataset. We applied a similar strategy for \textit{Deep-Lock}, and the validation accuracy over training iterations for all the models is shown in Figure~\ref{fig:fine_tuning}. We can observe from the figure that even if an adversary has access to 10\% of the original training data, he is not able to retrain the model with a comparable accuracy of the original model. In fact, the validation accuracies do not improve from a random classification.

\section{Conclusion}\label{sec:conclusion}
In this paper, we propose a lightweight, generic, key-based DNN IP protection scheme \textit{Deep-Lock} using an \textit{S-Box} and key-scheduling algorithms to defend against unauthorized usage of stolen DNN models. The method ensures that only an authorized user with a correct master key can accurately use the locked DNN model, and a wrong master key will provide random classification. \textit{Deep-Lock} does not modify any structural details of a DNN model, making it scalable for all existing software and hardware DNN implementations without adversely affecting performance. We evaluated \textit{Deep-Lock} for various DNN architectures and datasets. We have also demonstrated its robustness against model fine-tuning attack.

\bibliographystyle{ACM-Reference-Format}
\bibliography{reference}

\end{document}